\documentclass[conference]{IEEEtran}
\IEEEoverridecommandlockouts

\usepackage{cite}
\usepackage{amsmath,amssymb,amsfonts}
\usepackage{algorithmic}
\usepackage{graphicx}
\usepackage{textcomp}
\usepackage{xcolor}
\usepackage{outlines}
\usepackage[normalem]{ulem}
\usepackage{lipsum}
\usepackage{epsfig} % for postscript graphics files
\usepackage{subcaption}
\usepackage[hyphens,spaces,obeyspaces]{url}
\usepackage[colorlinks,allcolors=blue]{hyperref}
\usepackage{comment}
\usepackage{caption}
\usepackage{dsfont}
\graphicspath{figs/}
\definecolor{ao(english)}{rgb}{0.0, 0.5, 0.0}

\usepackage{tabularx,booktabs}
\newcolumntype{C}{>{\centering\arraybackslash}X} % centered version of "X" type
\usepackage{xspace}
\newcommand{\etal}{\mbox{\emph{et al.\ }}}

\usepackage{xargs}
\usepackage[colorinlistoftodos,prependcaption,textsize=tiny]{todonotes}
\usepackage[colorinlistoftodos,prependcaption,textsize=tiny]{todonotes}
\newcommandx{\add}[2][1=]{\todo[linecolor=red,backgroundcolor=red!25,bordercolor=red,#1]{#2}}

\def\BibTeX{{\rm B\kern-.05em{\sc i\kern-.025em b}\kern-.08em
    T\kern-.1667em\lower.7ex\hbox{E}\kern-.125emX}}

\makeatletter
\newcommand{\linebreakand}{%
  \end{@IEEEauthorhalign}
  \hfill\mbox{}\par
  \mbox{}\hfill\begin{@IEEEauthorhalign}
}
\makeatother

\begin{document}

\title{Handwashing Action Detection System for an Autonomous Social Robot\\
}
\author{
\IEEEauthorblockN{Sreejith Sasidharan}
\IEEEauthorblockA{\textit{AMMACHI labs} \\
\textit{Amrita Vishwa Vidyapeetham}\\
Amritapuri, India \\
0000-0002-9049-8759}
\\\hfill
\IEEEauthorblockN{Anand M Das}
\IEEEauthorblockA{\textit{Department of Mechanical Engineering} \\
\textit{Amrita Vishwa Vidyapeetham}\\
Amritapuri, India \\
mohandas.anand16@gmail.com}
\and
\IEEEauthorblockN{Pranav Prabha}
\IEEEauthorblockA{\textit{AMMACHI labs} \\
\textit{Amrita Vishwa Vidyapeetham}\\
Amritapuri, India \\
0000-0002-8547-5691}
\\\hfill
\IEEEauthorblockN{Chaitanya Kapoor}
\IEEEauthorblockA{\textit{Department of Computer Science} \\
\textit{Amrita Vishwa Vidyapeetham}\\
Amritapuri, India \\
kapoorchaitanya42@gmail.com}
\and
\IEEEauthorblockN{Devasena Pasupuleti}
\IEEEauthorblockA{\textit{Department of Mechanical Engineering} \\
\textit{Amrita Vishwa Vidyapeetham}\\
Amritapuri, India \\
0000-0003-3980-0219}
\\\hfill
\IEEEauthorblockN{Gayathri Manikutty}
\IEEEauthorblockA{\textit{AMMACHI labs} \\
\textit{Amrita Vishwa Vidyapeetham}\\
Amritapuri, India \\
0000-0003-2245-1550}

\linebreakand
\IEEEauthorblockN{Praveen Pankajakshan}
\IEEEauthorblockA{\textit{VP, Data Science \& AI} \\
\textit{Cropin AI Labs}\\
Bengaluru, India \\
0000-0001-9545-2971}
\and
\IEEEauthorblockN{Bhavani Rao}
\IEEEauthorblockA{\textit{AMMACHI labs} \\
\textit{Amrita Vishwa Vidyapeetham}\\
Amritapuri, India \\
0000-0003-2626-1973}
}

\maketitle

%%%%%%%%%%%%%%%%%%%%%%%%%%%%%%%%%%%%%%%%%%%%%%%%%%%%%%%%%%%%%%%%%%%%%%%%%
\begin{abstract}
Young children are at an increased risk of contracting contagious diseases such as COVID-19 due to improper hand hygiene. An autonomous social agent that observes children while handwashing and encourages good hand washing practices could provide an opportunity for handwashing behavior to become a habit. In this article, we present a human action recognition system, which is part of the vision system of a social robot platform, to assist children in developing a correct handwashing technique. A modified convolution neural network (CNN) architecture with Channel Spatial Attention Bilinear Pooling (CSAB) frame, with a VGG$-16$ architecture as the backbone is trained and validated on an augmented dataset. The modified architecture generalizes well with an accuracy of $90\%$ for the WHO-prescribed handwashing steps even in an unseen environment. Our findings indicate that the approach can recognize even subtle hand movements in the video and can be used for gesture detection and classification in social robotics.
\end{abstract}


%%%%%%%%%%%%%%%%%%%%%%%%%%%%%%%%%%%%%%%%%%%%%%%%%%%%%%%%%%%%%%%%%%%%%%%%%
\section{Introduction}

\label{sec:intro}
Poor hand hygiene practices among children make them vulnerable to morbidities such as anemia, respiratory illnesses, and diarrhea \cite{hussam2017habit}. The simple act of hand washing with soap can potentially prevent up to one million deaths worldwide \cite{Curtis2003}. When endowed with autonomous behavior, an agent that watches children while hand washing could provide behavioral nudges that can make hand washing a regular habit. Prior {\it Wizard of Oz} studies conducted by researchers with a social robot called Pepe showed significant improvement ($40\%$) in frequency and quality of children's hand washing practices pre and post-intervention. \cite{Unnikrishnan_2019, Deshmukh_2019}. 
\begin{figure}[thpb] 
    \centering
    \includegraphics[width=0.8\linewidth]{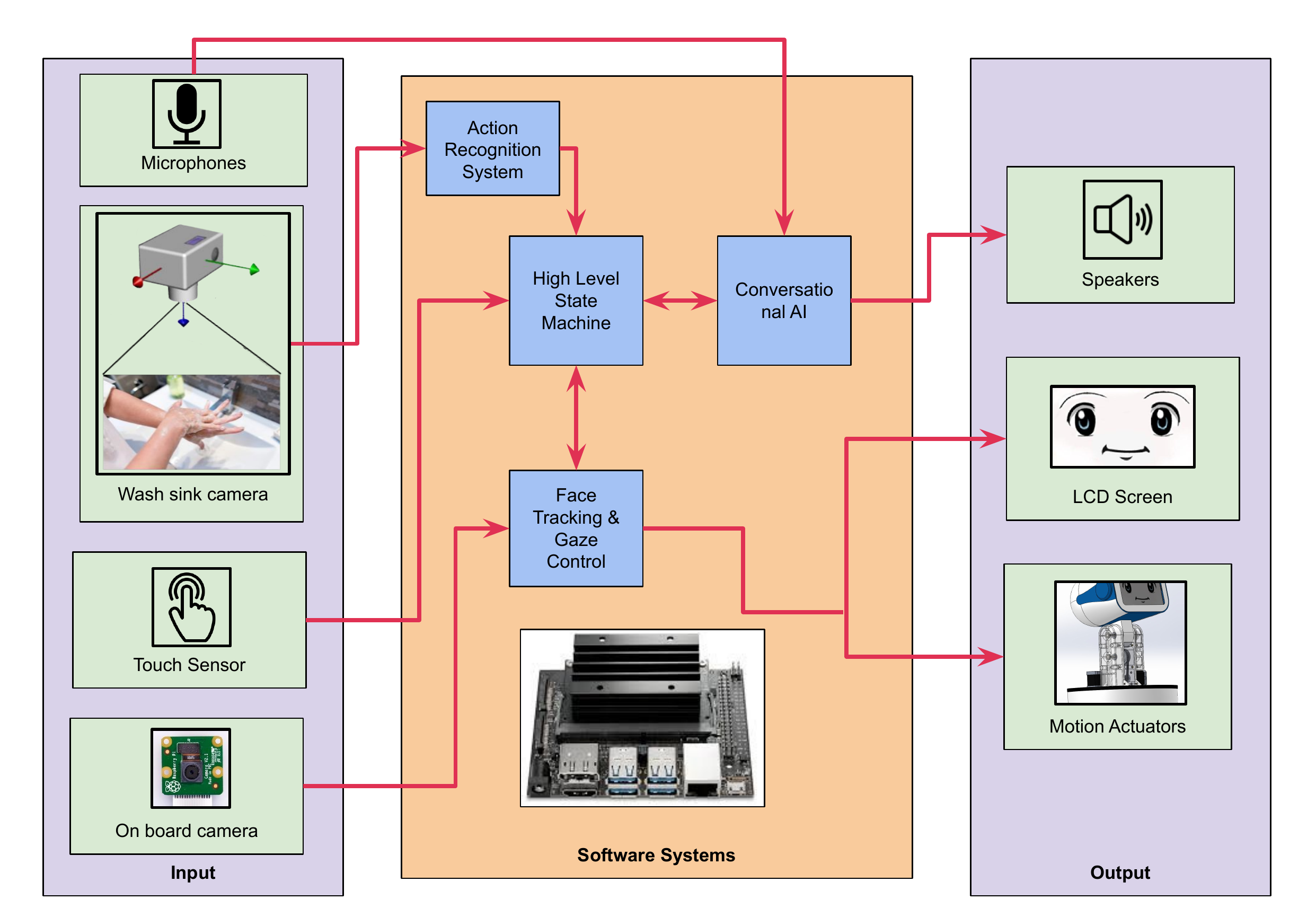}
    \caption{Architecture of the Social Robot showing also the action recognition system.}
    \label{fig_system_block}
\end{figure}

One of the essential considerations for designing an autonomous agent to promote good hand hygiene is a robust human action recognition (HAR) system for analyzing and understanding human action. For such a system to be effective, it needs to have high accuracy in classifications of human actions and also have a high adaptive ability. Our main objective in this research is to design a HAR system, for a social robot, whose vision system can detect the different hand washing steps prescribed by the World Health Organization (WHO)\cite{WHO_2017} and classify it accurately in order to monitor the quality of hand washing. Fig.~\ref{fig_system_block} shows the block diagram of our proposed system. We have focused our work on the six main steps of hand hygiene which are common to both hand washing with soap and hand rubbing with alcohol-based formulation, as prescribed by the WHO guidelines. 

We conducted a Human Robot Interaction (HRI) co-design study in 2021 using a participatory approach to design the embodiment of the social agent with children. Based on the quantitative and qualitative interview results, we have designed a prototype for the physical embodiment of the social robot \cite{iros}. We also developed a collaborative gaming platform where the children can play with the robot to learn the correct steps of handwashing \cite{roman}. To assess and provide the right feedback to children while handwashing in real-time, a robust handwashing action recognition vision model is required. We trained the robot prototype's vision model on an augmented dataset representing real-world environmental scenarios on hand washing. Real world scenarios demand the HAR systems for classification to work under large intra-class variations such as hand pose variations, skin color and size variations, and under varying environmental conditions, such as different lighting conditions, camera position variations, occlusions, shadow, blur, all of which pose challenges for robot's performance. Therefore a comprehensive training dataset that matches the real world scenarios as closely as possible is essential for the success of the social robot.

% HAR essentially consists of two parts, namely: Representation and
% Classification. Zhang \etal\cite{zhang2017review} reviewed various methods of
% representation based on  Optical flow, $3$D STVs (Spatio-Temporal Volumes), and skeleton-based representations. The methods of classification in HAR can be a top-down or a bottom-up approach and can be grouped into : template-based
% approach, generative models and discriminatory models approach
% \cite{zhang2017review}.

%%%%%%%%%%%%%%%%%%%%%%%%%%%%%%%%%%%%%%%%%%%%%%%%%%%%%%%%%%%%%%%%%%%%%

\section{Prior Work on HAR}
\label{sec:priorwork}

Early studies on conventional learning strategies discussed the lack of realistic and annotated datasets or using manually-built, local spatio-temporal feature points (in low dimensional spaces) to train the classifiers \cite{Laptev2008}. Such an approach becomes problem-centered and might not scale. Convolutional neural networks (CNN) can be used for HAR classification for HAR classification and representation when the data is diverse and rich \cite{zhang2017review}. Each hand washing action from an input video stream is fed as a sequence of images to the CNN, thereby creating a frame-based image classifier. The ``sequence'' and the order in which the frames are recognized, are important for providing feedback and for behavior corrections. 

Although CNN's have very high accuracy in image recognition classification problems, inorder to overcome the vanishing/exploding gradient \cite{hochreiter1998vanishing}, Residual network (ResNet) \cite{resnet} were introduced containing residual blocks. Cikel \etal conducted a study focusing on the classification of hand washing steps using ResNet with LSTMs \cite{kevin-mario}. The authors created two custom datasets - the RGB dataset - which featured information about the spatial video features and RGB frames, and the optical flow dataset - which contained information about the temporal video features, obtained by applying the Dual TV LI optical flow algorithm to the original dataset. Both these datasets were trained using a ResNet$-152$ model encoder combined with an LSTM decoder, pre-trained on the ImageNet dataset. The authors found that considering all the $12$ classes of the dataset; the RGB dataset offered the highest accuracy ($78.67\%$). Upon reducing the classes to seven, the accuracy of the RGB network significantly increased to $97.33\%$. 

Chen \etal \cite{chen2017sca} studied the effectiveness of the Spatial and Channel-wise Attention in Convolutional Neural Network (SCA-CNN) framework for image captioning. CNN models VGG-19 and ResNet-152 were used for image encoding part, and an LSTM model was used to generate caption words after decoding. The results show that SCA-CNN makes the original CNN multi-layer feature maps adaptive to the context-specific channel-wise attention and spatial attention at multiple layers. Another work that emphasizes the importance of attention layers was proposed by Woo \etal \cite{Woo}. Convolutional Block Attention Module (CBAM) is a straightforward but efficient attention module for feed-forward convolutional neural networks. The module sequentially infers attention maps along the two distinct dimensions of the channel and spatial from an intermediate feature map. It then multiplies the attention maps by the input feature map for adaptive feature refining. Their tests demonstrate the broad applicability of CBAM by showing consistent increases in classification and detection results with different models. 

In our approach to handwashing action recognition, after testing multiple architectures, we converged on the ResNet$-50$ architecture as the backbone. We trained it on an open-source handwashing dataset based on the WHO prescribed handwashing steps. However, the features extracted by the residual network were not discriminatory enough for accurately classifying the hand washing steps. To further improve the model's performance, we built and added an attention module on top of the CNN model. Adding the attention layers significantly improved the model's performance, and the details of the same are presented in the upcoming sections.

To summarize, the main contribution of this article to the new and evolving body of knowledge, HAR for HRI, include the presentation of:
\begin{itemize}
    \item Transfer learning based methodology to leverage pre-trained models to reduce the training data required,
    \item Modified architecture with a Channel Spatial Attention Bilinear Pooling (CSAB) module for classifying hand washing steps.
\end{itemize}

%%%%%%%%%%%%%%%%%%%%%%%%%%%%%%%%%%%%%%%%%%%%%%%%%%%%%%%%%%%%%%%%%%%%%%%%%%%%%%%%

\section{Background}
\label{sec:background}

\subsection{Backbone Architecture}
\label{ssec:arch}
A significant challenge posed by the handwashing step classification task is that the inter-class similarities are usually high. Similarly, intra-class differences also tend to be high due to the way different people perform the handwashing step. Such classification problems are called Fine-Grained Image Classifications. Wang \etal proposed a bilinear pooling theory that effectively utilizes both channel and spatial feature information to solve fine-grained classification problems \cite{wang_CSAB_2019}. We adapted the Channel Spatial Attention Bilinear Pooling (CSAB) frame from this work to solve the handwashing step classification problem. The complete model architecture we used is discussed below.

In fine-grained image classification problems, the key is to find differences in local areas in the image. Therefore, it is often challenging to use global features extracted by a single convolution layer for making predictions. An attention mechanism can be used to pay attention to relevant regions of the image and provide more weight to those regions.

\begin{figure}[hpbt]
\centerline{\includegraphics[width=0.5\textwidth]{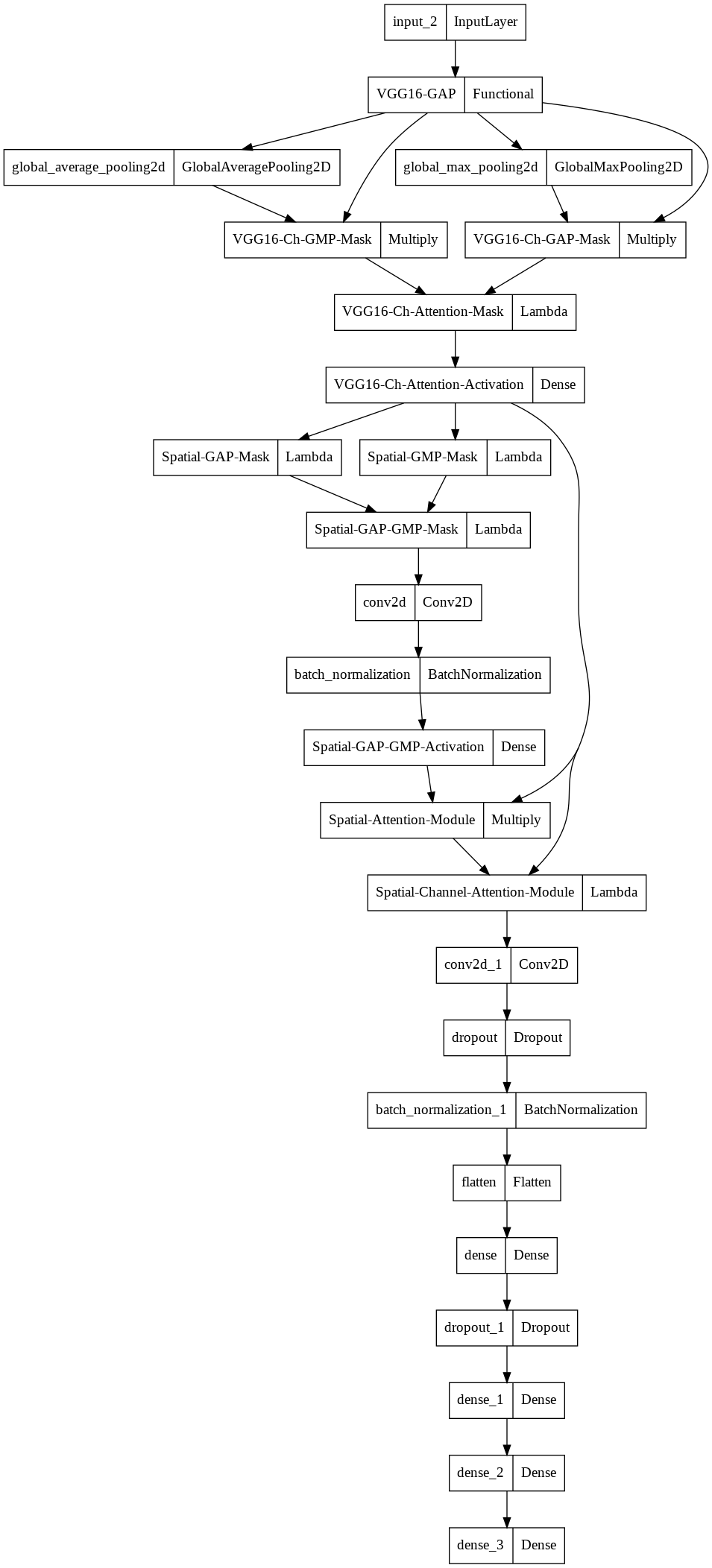}}
\caption{ Architecture with Channel Spatial Attention Bilinear Pooling  (CSAB) frame for recognizing the handwashing steps.}
\label{fig_csab_arch}
\end{figure}

A VGG-$16$ \cite{simonyan2015deep} model is taken as the backbone architecture and it is pre-trained on ImageNet. The CSAB model we have used combines channel attention and spatial attention simultaneously in a double attention module. In this dual attention model, attention in both spatial and channel dimensions complement each other. The overall architecture can be seen in Fig.~\ref{fig_csab_arch}. If we represent the input image as $\mathbf{X}\in \Re^{M\times N\times 3}$ and $\mathbf{y}$ as the output labels $\in\mathds{Z}^{+}$. By $\mathbf{Y}\in \Re^{H\times W\times C}$, we refer to the features generated prior to the fully convoluted layer with height $H$, width $W$ and channel $C$. The global feature context is given by the global average
pooling (GAP) as
\begin{equation}
g(\mathbf{Y}) = \frac{1}{H\times W}\sum_{i=1}^{H}\sum_{j=1}^{W}Y_{ij:}\\
\nonumber
\end{equation} and the global max pooling (GMP) is:
\begin{equation}
f(\mathbf{Y}) = \frac{1}{H\times W}\max_{i}\max_{j}Y_{ij:}\\
\nonumber
\end{equation}
The Channel Attention mask can then be denoted as: $\sigma(\mathbf{W}_{1}((g(\mathbf{Y})\cdot \mathbf{Y})+(f(\mathbf{Y})\cdot \mathbf{Y}) ))$, where $\cdot$ represents Hadamard Multiplication and $\mathbf{W}_{1}\in \Re^{1\times C}$ is the weights of the FCN and $\sigma$ is the sigmoid activation function.

% \textcolor{purple}{The first module of CSAB is the channel attention mechanism. This module helps in enhancing channel information that is significant and suppress insignificant channel information. We take feature, $F$, from the feature extractor and apply Global average pooling (GAP) and Global max pooling (GMP) on it to get $F_{MAP}$ and $F_{GMP}$ respectively. $F_{MAP}$ utilises the full spatial information , while $F_{GMP}$ indicates a channel's critical information. $F_{MAP}$ and $F_{GMP}$ are fused together. }

 \begin{figure*}[th]
    \captionsetup[subfigure]{justification=centering}
        \begin{subfigure}[b]{.33\textwidth}
            \centering
            \includegraphics[width=\linewidth]{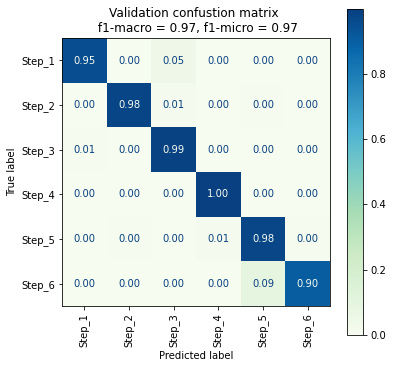}
            %\vspace{-2\baselineskip}
            \caption{Validation Confusion Matrix\\for $model\;1$}
            \label{fig:VGGSCA}
        \end{subfigure}
        \begin{subfigure}[b]{.33\textwidth}
            \centering
            \includegraphics[width=\linewidth]{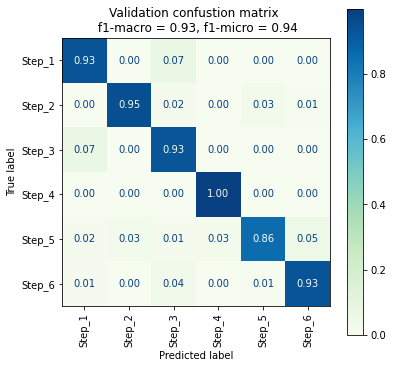}
            %\vspace{-2\baselineskip}
            \caption{ Validation Confusion Matrix\\for $model\;2$}
            \label{fig:RESNET_Plain}
        \end{subfigure}
        \begin{subfigure}[b]{.33\textwidth}
            \centering
            \includegraphics[width=\linewidth]{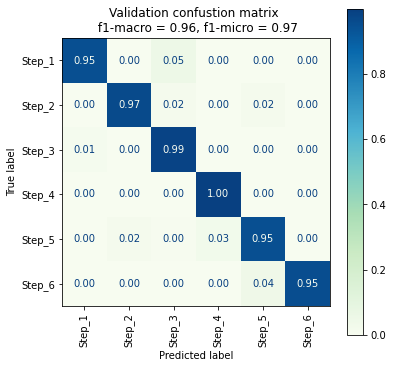}
            %\vspace{-2\baselineskip}
            \caption{Validation Confusion Matrix\\for $model\;3$}
            \label{fig:VGG_Plain}
        \end{subfigure}
        \caption{Confusion Matrices}
        \label{fig_cf}
    \end{figure*}

\subsection{Dataset}

For a model to work efficiently in real-life scenarios, it has to be
trained with input data collected under different conditions and varying environments.

\begin{figure}[ht]
\centering
\includegraphics[width=\linewidth]{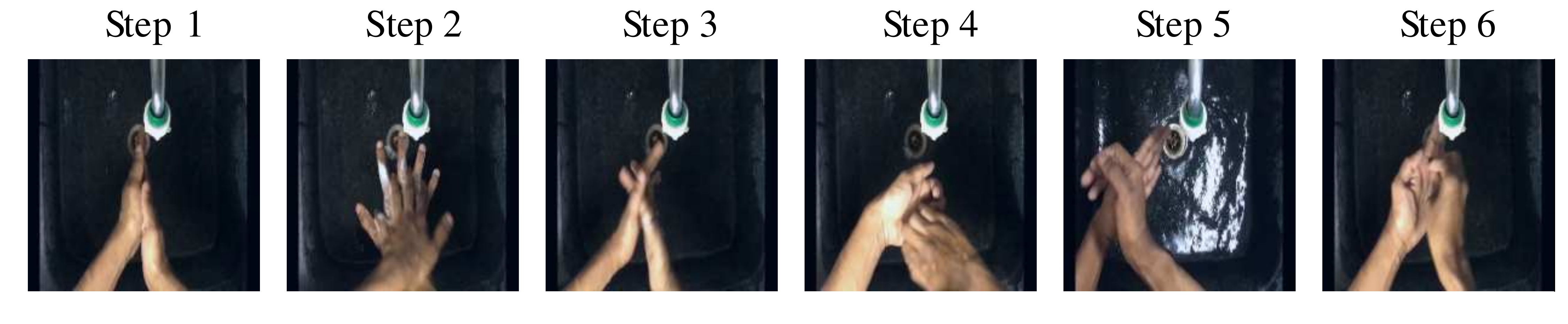}
\caption{Sample frames from the dataset illustrating the Handwashing steps recommended by WHO}
\label{fig_dataset_classes}
\end{figure}

For our study, we chose an open-source dataset of WHO handwashing steps \cite{kaggle_handwash_dataset_2019}, which provides $292$
samples for the $12$ steps in the handwashing (see Fig.~\ref{fig_dataset_classes}). This
publicly available dataset is recorded against five different sink
backgrounds. This video is most suitable for applications in the real world as it has
samples with varying parameters like illumination, field of view, and
background. 

%%%%%%%%%%%%%%%%%%%%%%%%%%%%%%%%%%%%%%%%%%%%%%%%%%%%%%%%%%%%%%%%%%%%%%%%%%%%%%%%

\section{Methodology}

\subsection{Data Preparation}
\label{ssec:data_prep}
The video dataset had to be converted into image sequences so that it could be used to train a deep neural network for image classification. For this, we extracted individual frames from the video samples using OpenCV. We manually chose representational images that are visually similar to the six steps of handwashing prescribed by the World Health Organisation. We also excluded images with severe motion blur and images without any hands in the frame. The left-hand and right-hand versions of the handwashing steps two, four, five, and six were further combined to form six classes in the dataset, representing the six steps of handwashing. 

We further modified the dataset with an environment-level train-test split to test the generalizability of the trained model in unfamiliar environments. We did this by using four different background environments in the dataset for training the model. The fifth environment, which the model did not see during training, was used for testing the model's performance.

%%%%%%%%%%%%%%%%%%COMMENTED SECTION%%%%%%%%%%%%%%%%%%%%%%
\begin{comment}
During training, the model starts to recognise the frequently occurring features in 
the training dataset and can easily learn unwanted features such as repeating 
background of sinkholes or taps instead of learning the hand features.
This issue can be observed in the saliency maps shown in Fig.~\ref{fig_saliency_unwanted}. 
Salie ncy maps are images that depict the features of an image that the neural network is
'looking at' to make a prediction. In the following section, we dwell deeper into the 
techniques that we used to solve this issue.
\end{comment}
%%%%%%%%%%%%%%%%%%%%%%%%%%%%%%%%%%%%%%%%%%%%%%%%%%%%%%%%%

\subsection{Data Augmentation}

Data Augmentation \cite{cshorten2019} is a fundamental technique for obtaining high performance for fixed neural network architecture. Two primary reasons to augment the data are to improve accuracy and generalization. While the accuracy can be improved by adding more training data, the generalization ability of these models is a difficult challenge. By generalizability, we refer to the ability of the model to not lose its performance when evaluated on unseen data. Since there is also a limited variety of environmental conditions in the training data, data augmentation was utilized to increase the diversity of the training data. The following transformations were done on the images to create an augmented dataset.

\begin{figure}[h]
\centering
\includegraphics[width=\linewidth]{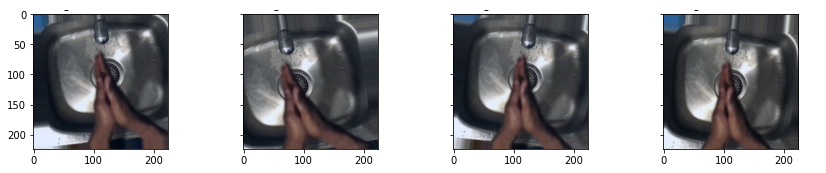}
\caption{Image samples from the augmented dataset}
\label{Augmented Dataset}
\end{figure}

\subsubsection{Geometric Transformations}
Geometrical transformations can remove any positional bias of the hands. We used random rotations, width shifts, random scaling, and horizontal flips to minimize positional biases.  
    
\subsubsection{Color Space Transforms}

The training image frames were augmented by adding spatial components and color to the background. For color augmentation, we used a custom pre-process function, adding random contrast and brightness values over 70\% to 130\% of the original pixel values of the image to image batches before training. This simulated real-world change in the lighting conditions and the foreground object.

\begin{table*}
 \caption{Comparison of performance of various models}
\label{table:comparison}
\begin{tabularx}{\textwidth}{@{}l*{10}{C}c@{}}
No. &Attention module &Backbone    &Epoch &Batch Size &Train acc. &Validation acc. &Test acc. &Custom data-1 &Custom data-2 &Custom data-3\\\midrule
model-1       &Yes             &VGG-16                 &10    &128    &0.99     &0.97   &0.90  &0.61  &0.83   &0.64        \\
model-2       &No              &ResNet-50            &10    &128    &0.99     &0.94     &0.85  &0.74  &0.62   &0.58        \\
model-3       &No              &VGG-16                &10    &128    &0.99     &0.97    &0.89  &0.74  &0.77   &0.62        \\
\end{tabularx}
\end{table*}

The models were developed with Keras ($>2.4$) built on top of Tensorflow $2.0$
and was trained on a machine with $25$GB RAM and an NVIDIA$^\text{\textregistered}$ Tesla P$100$ GPU. 

After extensive hyper-parameter tuning, one optimal model was selected from each of the three different neural network architectures used for training and is presented in this paper. Apart from the model with the CSAB module, which has a VGG-16 backbone, a ResNet$-50$ as well as a VGG-16 based models, with just the backbone architecture that outputs into a fully connected, dropout, and a final dense layers with softmax activation were also trained using the same dataset. This was done to evaluate the effects of the added attention layer. 

% \begin{figure}[b]
%     \centering
%     \includegraphics[width=.4\linewidth]{figs/hist_acc.eps}
%     \label{fig:sfig_epoch_acc}
%     % \hspace{2}
%     \includegraphics[width=.4\linewidth]{figs/hist_loss.eps}
%     \label{fig:sfig_epoch_loss}
%     \caption{(a) Epoch Accuracy (b) Epoch Loss}
% \end{figure}

%%%%%%%%%%%%%%%%%%%%%%%%%%%%%%%%%%%%%%%%%%%%%%%%%%%%%%%%%%%%%%%%%%%%%%%%%%%%%%%%

\section{Results}
\label{sec:results}
    
\begin{figure*}[h]
% \begin{figure}[thpb]
\centering 
\includegraphics[width=0.9\linewidth]{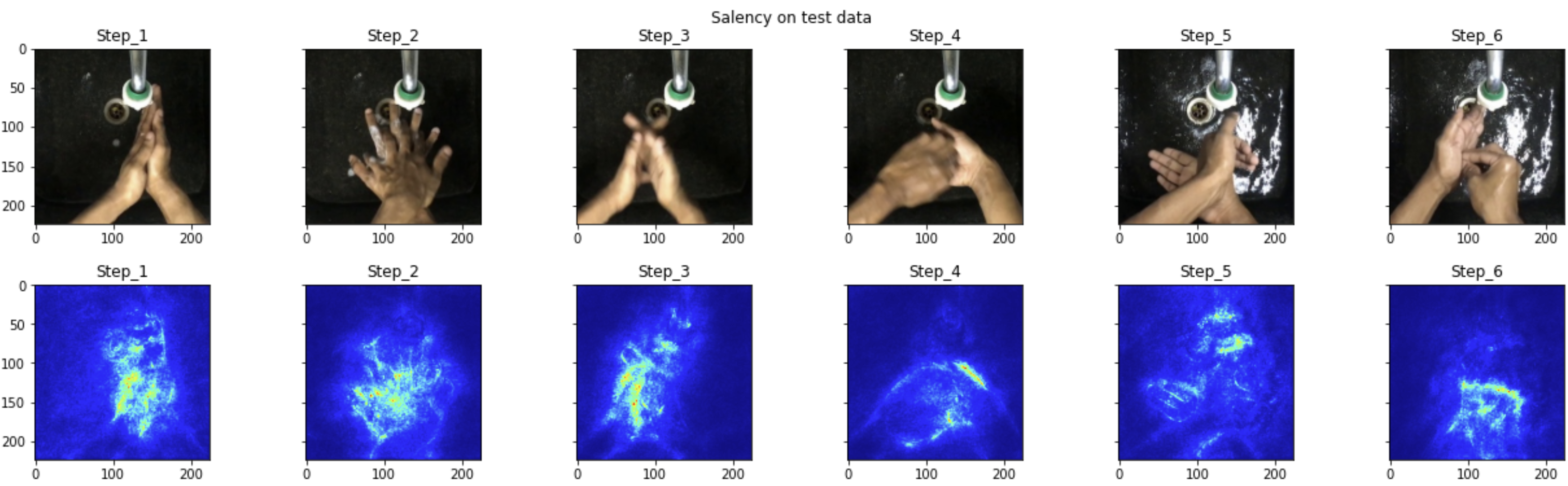}
\caption{Saliency map of the model-$1$ on the test data}
\label{fig_attention_test}

\includegraphics[width=0.9\linewidth]{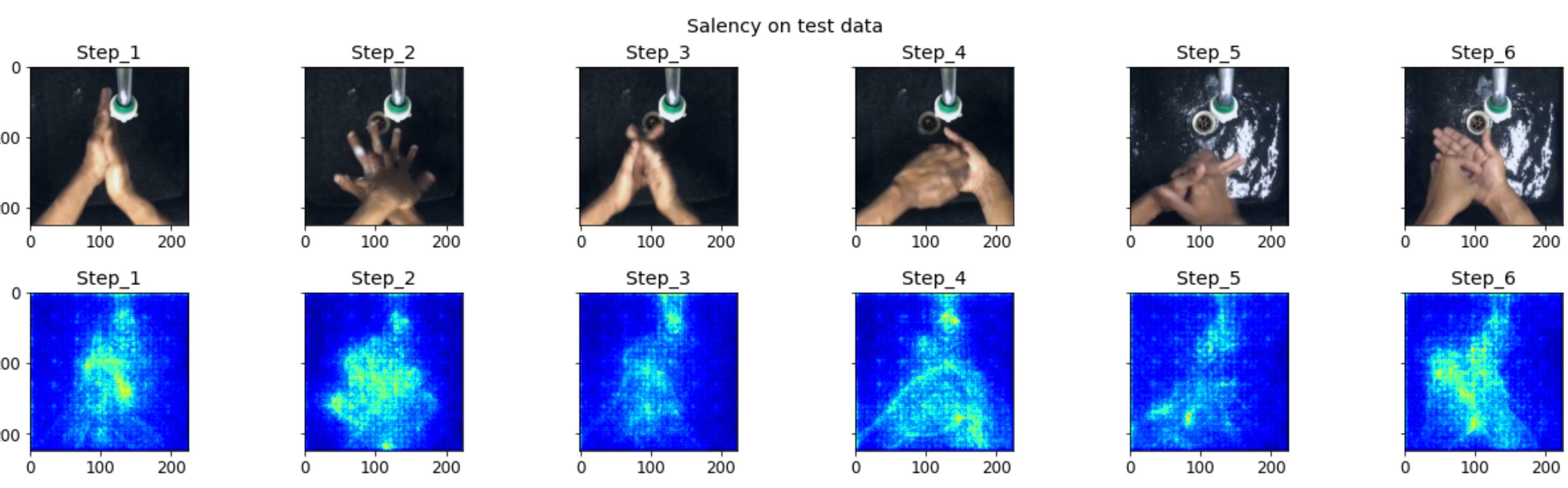}
\caption{Saliency map of the model-$2$ on the test data}
\label{fig_resnet_test}

\end{figure*}
%Training time:
%The training time was $7.5$ hours for the model-$1$  which had the self-gated attention module, and $3.95$ hours
%for model $5$, without the self-gated attention module.  

%Effect of data augmentation:
%Fig.~\ref{fig_attention_test} shows the saliency map of model-$1$. Compared to Fig.~\ref{fig_saliency_unwanted} the attention of this model is more on the hand which is the feature of interest than on the background features. This clearly shows the positive effect of data augmentation of the dataset on the test results. 

Generalizability: A detailed comparison of the performance of different models can be found in Table~\ref{table:comparison}. The highest test accuracies were observed for models with a CSAB module ($1$). While the accuracies on the Kaggle dataset are the same between model 1 and model 3, model 1 is more generalizable on custom datasets, as shown in Table~\ref{table:comparison}. The three custom datasets were created by extracting image frames from real-time video data of three different persons performing the six steps of handwashing. This data was not part of the original kaggle dataset used for training, validation and testing of the models.

Inference time: For real-time applications inference time of a model is an important consideration. All models exhibited very fast inference times of $<$5ms.

It can be seen that the addition of CSAB module on top of the VGG-16 architecture increased test accuracy from 89\% to 97\%. The performance of the model is also better than the residual neural network architecture, which could produce a test accuracy of 91\%. The model could effectively distinguish the different steps of hand washing.

% \subsection{Experiments with Deep Learning Architectures}
% We have tested the dataset in various architectures, they include:
% \begin{itemize}
%     \item{Transfer Learning with VGG16 -- }
%     \item{Transfer Learning with ResNet-$50$ -- }
% \end{itemize}

The handwashing recognition system making inference on video can be seen in action \href{https://drive.google.com/file/d/1JFiO8c0VaFqA6djNpOv6tX-DvpQoqMYN/view?usp=sharing}{here}. The scripts and notebooks developed for this work are available for reproducibility\footnote{\url{https://github.com/voiD-96/HAR_Tencon2022.git}}. The dataset used for training the deep learning models is also available\footnote{\url{http://ieee-dataport.org/9903}}.
%\add{Check if the video link is correct}
% The scripts and notebooks developed for this work is available for reproducibility\footnote{\url{https://codeocean.com/[anonymized-for-review]}}.
%\subsection{Video Results and Reproducibility}
%\label{ssec:video}
%The handwashing recognition system making inference on video can be seen in action \href{https://drive.google.com/file/d/1fn7rfm-NX2Z2rYk-HDRV3t\_iqmYjz2Cp/view}{here}.
%\add{The video shows the 7 steps than the 6 steps defined in the article. This can be changed later.}

%%%%%%%%%%%%%%%%%%%%%%%%%%%%%%%%%%%%%%%%%%%%%%%%%%%%%%%%%%%%%%%%%%%%%%%%%%%%%%%%
\section{Conclusions and Future Work}

Our results indicate that the modified CNN network with a CSAB module for classification of the six steps of hand washing as prescribed by WHO has good generalizability with training accuracy of $99\%$ and validation accuracy of $97\%$. Enhancing the open-source augmented dataset with a richer variety of video frames with lesser inter-class similarities and including videos from a broader range of demographics such as age groups, gender, and ethnicity will improve the test accuracy of the model further. The authors would like to acknowledge that despite having many advantages and appearing to be a natural extension for image classification applications, deep learning HAR is still an evolving area of research for real-world scenarios. Many challenges remain due to high levels of inter-class similarities, intra-class variations, difficulty in capturing extended context, and a lack of standard benchmarking datasets.

Looking at the results from an HRI perspective, the HAR model designed as part of this study has established the feasibility of using deep neural networks in identifying hand washing steps with sufficiently good accuracies for practical real-world use in a hand washing social robot. When deploying this model in a real robot, the predictions made by the HAR system will be on a series of frames with temporal relationships, and hence hand washing step prediction accuracy can be further improved by fusing the temporal data with the predicted data made by the HAR system.  

As a part of future work, the authors believe that designing multi-stream CNNs that can learn spatial and temporal aspects of the hand washing activity can improve the performance of the hand washing recognition system. The model can be enhanced even further in the future by training it on datasets with much more variations in skin tones, hand washing style, and others. The performance of the CSAB model can further be improved by experimenting with different hyperparameter tuning methods. We intend to deploy the modified and trained model discussed in this article on our prototype to tailor the robot for hand hygiene education in schools. The proposed HAR system can also be used standalone in other application domains, including healthcare and food industries, to monitor hand washing compliance.

% \addtolength{\textheight}{-3cm}   % This command serves to balance the column lengths

%%%%%%%%%%%%%%%%%%%%%%%%%%%%%%%%%%%%%%%%%%%%%%%%%%%%%%%%%%%%%%%%%%%%%%%%%%%%%%%%
\section{Acknowledgments}
The authors express their gratitude to Sri Mata Amritanandamayi Devi, Chancellor, Amrita Vishwa Vidyapeetham, without whose constant guidance, support, and encouragement this project would not have been possible. We also extend our gratitude to our colleague and fellow researcher Tzur Sayag for his insights and support.

%%%%%%%%%%%%%%%%%%%%%%%%%%%%%%%%%%%%%%%%%%%%%%%%%%%%%%%%%%%%%%%%%%%%%%%%%%%%%%%%

%References are important to the reader; therefore, each citation must be complete and correct. If at all possible, references should be commonly available publications.

{\small
\bibliographystyle{IEEEtran}
\bibliography{egbib}
}

\end{document}